\documentclass[runningheads]{llncs}
\usepackage{graphicx}
\usepackage{amsmath,amssymb} 
\usepackage{color}

\usepackage{booktabs, multirow, bigstrut}

\newcommand\IMAGES{\mathcal{I}}
\DeclareMathOperator\COL{:}
\DeclareMathOperator\ra{\rightarrow}
\newcommand\set[1]{\{{#1}\}}

\newcommand\LABELS{\mathcal{L}}

\newcommand\EXPT{\mathbb{E}}
\newcommand\REALS{\mathbb{R}}

\DeclareMathOperator\ELEMWISE{\odot}

\newif\iffull
\fulltrue

\usepackage[linesnumbered,ruled,vlined]{algorithm2e}
\SetKwInput{KwInput}{Input}                
\SetKwInput{KwOutput}{Output}              

\iffull
\else
\usepackage{xr}
\fi

\begin{document}
\pagestyle{headings}
\mainmatter

\def\ACCV20SubNumber{762}  

\title{Visualizing Color-wise Saliency of\\ Black-Box Image Classification Models} 
\titlerunning{Visualizing Color-wise Saliency of Black-Box Image Classification Models}
\author{Yuhki Hatakeyama\inst{1*} \and
Hiroki Sakuma\inst{1} \and
Yoshinori Konishi\inst{1} \and
Kohei Suenaga\inst{2}}
\authorrunning{Y. Hatakeyama et al.}
%

\institute{SenseTime Japan, 4F, Oike Koto Building,
324 Oikeno-cho, Nakagyo-ku, Kyoto, Japan \\
\email{\{hatakeyama,sakuma,konishi\}@sensetime.jp}\\
Graduate School of Informatics, Kyoto University, 36-1 Yoshida-Honmachi,
Sakyo-ku, Kyoto, Japan \\
\email{ksuenaga@gmail.com} \\
}

\maketitle

\begin{abstract}
  Image classification based on machine learning is being commonly used.
  However, a classification result given by an advanced method, including deep learning, is often hard to interpret.
  This problem of interpretability is one of the major obstacles in deploying a trained model in safety-critical systems.
  Several techniques have been proposed to address this problem;
  one of which is RISE, which explains a classification result by a heatmap, called a \emph{saliency map}, that explains the significance of each pixel.
  We propose \emph{MC-RISE} (\textbf{M}ulti-\textbf{C}olor RISE), which is an enhancement of RISE to take color information into account in an explanation.
  Our method not only shows the saliency of each pixel in a given image as the original RISE does, but the significance of \emph{color components} of each pixel;
  a saliency map with color information is useful especially in the domain where the color information matters (e.g., traffic-sign recognition).
  We implemented MC-RISE and evaluate them using two datasets (GTSRB and ImageNet) to demonstrate the effectiveness of our methods in comparison with existing techniques for interpreting image classification results.
\end{abstract}

\section{Introduction}

\label{sec:intro}

As machine learning is widely applied to image classification,
there is a surging demand for the methods to explain classification results and
visualize it.  One can use such an explanation to check whether
a trained model classifies images based on a rational and acceptable criterion,
by which he or she can convince various stakeholders that the model is readily deployed.

\begin{figure}[tb]
  \centering
  \includegraphics[keepaspectratio, width=0.60\linewidth]{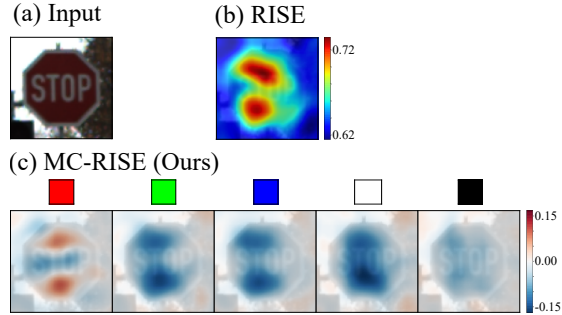}
  \caption{Visual comparison between the saliency map of  RISE~\cite{DBLP:conf/bmvc/PetsiukDS18} and the color sensitivity maps generated by our proposed method (MC-RISE).
  Best viewed in color.}
  \label{fig:example_maps}
\end{figure}

One of the most popular ways of the visualization is by
a \emph{saliency map}---a heatmap overlayed on the original image that indicates
which part of the image contributes to the classification result~\cite{lime,NIPS2017_7062,DBLP:conf/iccv/SelvarajuCDVPB17,DBLP:conf/bmvc/PetsiukDS18,selvaraju_grad-cam_2017,chattopadhay_grad-cam_2018,sundararajan_axiomatic_2017,smilkov_smoothgrad_2017,zeiler_visualizing_2014,springenberg_striving_2015,bach_pixel-wise_2015,shrikumar_learning_2017,zhang_top-down_2018,vasu_iterative_2020}.
Fig.~\ref{fig:example_maps} shows an example of a saliency map generated by a method called RISE~\cite{DBLP:conf/bmvc/PetsiukDS18}.
Given the image (a) of a road sign ``STOP'' and a model that indeed classifies this image as a stop sign,
RISE generates the saliency map (b), which indicates the part of (a) that contributes to the classification result by a heatmap.
From this saliency map, we can figure out that the part of
the sign surrounding the text ``STOP'' contributes positively to the classification result.

Among the explanation methods proposed so far, \emph{model-agnostic} techniques such as
LIME~\cite{lime}, SHAP~\cite{NIPS2017_7062}, and RISE~\cite{DBLP:conf/bmvc/PetsiukDS18} generate
a saliency map without accessing the internal information of a model, treating it as a black box.
Although these procedures differ in their detail, they all share the following central idea:
They compute a saliency map for a classification result by perturbing the given image and
observing how the output of the perturbed image changes from the original.
Concretely, given an image $I$, a model $M$, and its classification result,
they compute a saliency map by (1) generating perturbed image $I_1,\dots,I_N$ from $I$ by masking a part of it,
(2) computing the classification result $M(I_1),\dots,M(I_N)$ for each perturbed image, and
(3) comparing $M(I)$ with each $M(I_i)$.  If the output of a perturbed image with a specific part unmasked
tends to be the same as $M(I)$, then this part is considered to be important.

Although these model-agnostic explanation techniques give valuable insights about
a classification result of an image, there is important information
that has been overlooked by these methods: \emph{colors}.
They compute a saliency map indicating each pixel's existence,
but they do not take the color information of a pixel into account.
A \emph{color-aware} explanation is effective especially in the application domains
in which the colors in an image convey important information (e.g., traffic-sign recognition).

This paper proposes an extension of RISE~\cite{DBLP:conf/bmvc/PetsiukDS18}
so that it generates a color-aware saliency map.
We extend the original RISE so that it can compute a saliency map for each color of
a given color set; we call our extended method \emph{MC-RISE (\textbf{M}ulti-\textbf{C}olor RISE)}.
Fig.~\ref{fig:example_maps} (c) shows the saliency maps generated by MC-RISE.
It consists of five heatmaps, each of which corresponds to the significance of
a certain color of each pixel; the associated color is shown above each heatmap.
We can figure out that the red color at the peripheral part of
the stop sign contributes much to the classification result, whereas
the other colors do not contribute as much as red.

We also propose an enhancement to \emph{debias} the original RISE.
In our extension, the saliency of the pixels in an input image is close to $0$
if it is irrelevant for the classification result.
Notice that the heatmaps in Fig.~\ref{fig:example_maps} (c)
have value close to $0$ for the irreverent pixels, whereas the heatmap generated by
the original RISE in Fig.~\ref{fig:example_maps} (b) does not
contain a pixel of value $0$; it is hard to figure out
which part of the image is irreverent to the classification result
from the latter saliency map.

The contributions of the paper are summarized as follows.
(i) We propose a new model-agnostic explanation method for an image classifier, \emph{MC-RISE},
which generates \emph{color-aware} saliency maps for the classifier's decision.
(ii) To improve the interpretability of the saliency map, we propose a method to debias the saliency map of RISE
and incorporate it into MC-RISE.
(iii) We qualitatively and quantitatively compare our method to existing model-agnostic explanation methods
(LIME~\cite{lime} and RISE~\cite{DBLP:conf/bmvc/PetsiukDS18}) in GTSRB~\cite{gtsrb} dataset and
ImageNet~\cite{imagenet} dataset,
and showed that our method can extract additional information which can not be obtained by the existing methods, such as color-sensitivity.

\section{Related Work}
\label{sec:relatedWork}

Explanation-generating methods for an image classifier proposed so far can be categorized into the following two groups:
(1) ones that treat a model as a \emph{black box}
and
(2) ones that treat a model as a \emph{white box}.
The former observes the input--output relation in generating an explanation without using
the internal information of a model; whereas the methods in the latter group assume that the internal information is accessible.
As far as we know, our method is the first explanation-generating method for black-box models
that considers color information in generating an explanation.

Many white-box methods compute a saliency map utilizing the classifier's gradient information.
Grad-CAM~\cite{selvaraju_grad-cam_2017} and its extension  Grad-CAM++~\cite{chattopadhay_grad-cam_2018} use the gradient with respect to an intermediate feature map
to obtain class-specific weights for the feature map.
Integrated gradient~\cite{sundararajan_axiomatic_2017} and SmoothGrad~\cite{smilkov_smoothgrad_2017} accumulate the gradients with respect to modified input images in order to get a more interpretable sensitivity map than a single gradient map.
In \cite{zeiler_visualizing_2014,springenberg_striving_2015,bach_pixel-wise_2015,shrikumar_learning_2017,zhang_top-down_2018},
the relevance map for an intermediate layer is back-propagated layer-by-layer from the output layer to the input layer with a modified back-propagation rule.

However, it is pointed out that the explanation generated by gradient-based methods are not necessarily faithful to the classifier's decision process.
Adebayo et al.~\cite{adebayo_sanity_2018} shows that some gradient-based methods are nearly independent of the classifier's weight, and act like a model-ignorant edge detector rather than an explanation of the classifier.
\cite{heo_fooling_2019,dombrowski_explanations_2019,subramanya_fooling_2019} adversarially attack gradient-based methods and can manipulate a saliency map without regard to the classifier's output.
From these results, we expect that the methods based on the input--output relation are more faithful to the classifier's actual behavior than gradient-based methods.

Other white-box methods include optimization-based methods and attention-based methods.
In Meaningful perturbation~\cite{fong_interpretable_2017}, Extremal Perturbations~\cite{torchray}, and FGVis~\cite{wagner_interpretable_2019}, the image region to add perturbation (e.g. blurring, masking) are optimized by gradient descent with respect to an input image, and saliency information is extracted from the perturbed region.
In \cite{xu_show_2015}, a visual question answering model with the attention mechanism is proposed and the attention maps can be interpreted as the relevant parts in an image.

LIME (Local Interpretable Model-agnostic Explanation)~\cite{lime} and SHAP~\cite{NIPS2017_7062} are
popular explanation-generating methods for a black-box image-classifier.
Both perturb the given image preprocessed into a set of superpixels, observe how the output to the perturbed image,
and generate an explanation based on the change in the output.
Although their merits are widely appreciated, it is known that
a generated explanation is not robust to how an image is decomposed into superpixels~\cite{DBLP:journals/corr/abs-1910-07856}.
Our extension proposed in this paper is based on RISE~\cite{DBLP:conf/bmvc/PetsiukDS18} instead,
which does not require a prior preprocessing of an input image.
We will explain RISE in detail in \S\ref{sec:rise}.

Some black-box methods are based on these black-box techniques.
IASSA~\cite{vasu_iterative_2020} is an extension of RISE, which refines
a saliency map by iteratively adapting the mask sampling process based on the previous saliency map and attention map.
As extensions of the LIME framework, LORE~\cite{guidotti_local_2018} generates sample data with a genetic algorithm and fits decision trees instead of the linear regression model in LIME, and \cite{tsang_feature_2019} incorporates the effect of higher-order interactions between input features.

Other black-box methods include Anchors~\cite{ribeiro_anchors_2018}, which
searches for the minimal feature set which is sufficient for a correct prediction,
and CXPlain~\cite{schwab_cxplain_2019}, which trains the causal explanation model
for the classifier's behavior when some input feature is removed.

\section{RISE}
\label{sec:rise}

This section explains an explanation-generating method RISE~\cite{DBLP:conf/bmvc/PetsiukDS18},
which is the basis of our method.  For a detailed exposition,
see Petsiuk et al.~\cite{DBLP:conf/bmvc/PetsiukDS18}.

We first designate several definitions to define RISE.
An \emph{image} is a mapping from a finite set $\Lambda$ of \emph{pixels} to $\REALS^3$.
For an image $i$ and a pixel $\lambda$, the tuple $i(\lambda) \in \REALS^3$ is the RGB value of $\lambda$ in $i$.
We write $\IMAGES = \{\, i \mid i: \Lambda \ra \REALS^3 \,\}$ for the set of images.
We also designate the finite set of labels $\LABELS$.
We fix an image-classification model $M \COL \IMAGES \times \LABELS \ra [0,1]$ throughout this paper;
$M(i,l) \in [0,1]$ is the model confidence in classifying $i$ as $l$.
For any $i$, we assume that $\sum_{l \in \LABELS} M(i,l) = 1$.

A \emph{mask} is an element of $\Lambda \ra \set{0,1}$.
A mask $m$ represents an image transformation that sets the RGB value of a pixel $\lambda$
to $(0,0,0)$ if $m(\lambda)=0$; the transformation keeps the original value of $\lambda$ if $m(\lambda)=1$
Therefore, the image $i \ELEMWISE m$ that is obtained by applying the image transformation $m$ to an image $i$
is defined by $\lambda \mapsto i(\lambda) \times m(\lambda)$.

Given an image $i \in \IMAGES$ and a label $l \in \LABELS$,
we define the \emph{saliency map} $S_{i,l}$ as
$\lambda \mapsto \EXPT_{m}[M(i \ELEMWISE m, l) \mid m(\lambda) = 1]$, where the expectation is taken
over all the possible masks.
The intuition behind this definition is that $\lambda$ can be considered to positively contribute to
the classification of $i$ to $l$ if the model confidence of classifying the image to $l$
remains high when $\lambda$ \emph{is not masked.}

Let $X$ be a random variable whose value is a mask.
Then, the definition of $S_{i,l}(\lambda)$ is equal to
\begin{equation}
  \frac{1}{P[X(\lambda)=1]} \sum_{m} M(i \ELEMWISE m, l) \times m(\lambda) \times P[X=m] ~. \label{eq:riseFormula}
\end{equation}
This expression shows that we can compute $S_{i,l}(\lambda)$ by
computing the model confidence $M(i \ELEMWISE m, l)$ for every mask $m$
that does not mask $\lambda$ and taking the weighted sum of these values.

\begin{remark}
  \label{remark:optimization}
In the actual implementation of RISE, the following optimizations are often applied.
\begin{itemize}
  \item It is prohibitively expensive to precisely computing the value of the expression (\ref{eq:riseFormula}).  Therefore, $S_{i,l}$ is approximated by the Monte-Carlo method.
  We randomly sample sufficiently many masks $m_1,\dots,m_N$ and compute
  \begin{eqnarray}
    \label{eq:rise_deriv}
    S_{i,l}(\lambda) & \approx & \frac{1}{N} \sum_n \frac{m_n(\lambda)}{p} M(i \odot m_n, l) ~.
  \end{eqnarray}
  Here, $p$ is $P[X(\lambda)=1]$, $m_n$ is the $n$-th mask,
  and $N$ is the number of samples.

  \item RISE preprocesses a mask by bilinear interpolation before it applies the mask to an image.
  This is based on the intuition that (1) the saliency of a pixel is not binary and
  (2) a pixel that is close to an important pixel is often important.  After this optimization,
  a mask is an element of $\Lambda \rightarrow [0,1]$ instead of $\Lambda \rightarrow \set{0,1}$.
\end{itemize}
Our extensions proposed in this paper also uses these optimizations.
\end{remark}

\section{Debiased RISE and MC-RISE}

\begin{figure}[tb]
  \centering
  \includegraphics[keepaspectratio, width=\linewidth]{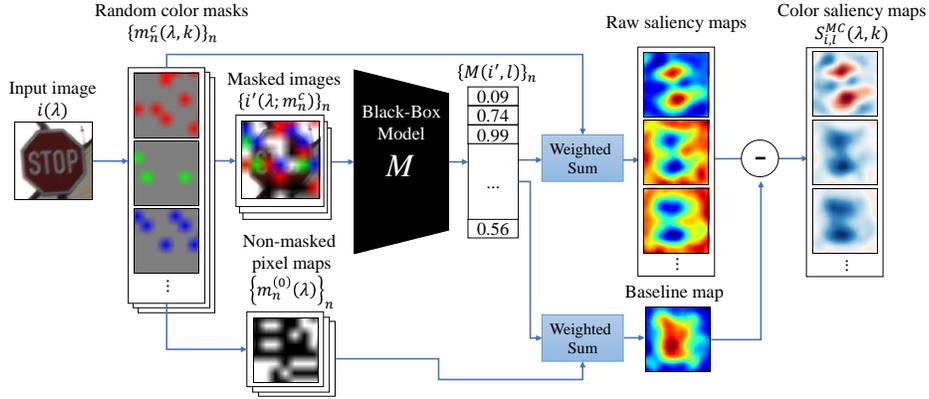}
  \caption{Overview of our proposed method MC-RISE.
  An input image $i(\lambda)$ is randomly masked by given colors and the masked images $i'(\lambda; m^{\mathrm{c}}_n)$ are fed to the black-box model $M$. Raw saliency maps are obtained by the weighted sum of the color masks $m^{\mathrm{c}}_n(\lambda,k)$, where the weights are the output probabilities $M(i',l)$. Then a baseline map obtained from non-masked pixel maps $m^{(0)}_n(\lambda)$ is subtracted from them so that the saliency value in image parts irrelevant to classification is aligned to zero.
}
  \label{fig:method}
\end{figure}


This section describes our extension to RISE.
As we mentioned in \S\ref{sec:intro}, our extension consists of two enhancements of RISE:
(1) MC-RISE, which generates \emph{color-aware} saliency maps, and
(2) \emph{debiasing} saliency maps, which is a tweak to the RISE procedure so that the value of an irrelevant pixel in a saliency map is close to $0$.
We first present (2) in \S\ref{sec:method_debias} and then (1) in \S\ref{sec:method_mcrise}.

\subsection{Removing Bias from Saliency Maps}
\label{sec:method_debias}

The range of the saliency values of a heatmap generated by RISE
is highly variable depending on an input image.
For example, the values in the saliency map at the first row in Fig.~\ref{fig:mcrise_gtsrb} (c)
ranges from $0.016$ to $0.023$, whereas the one at the third row ranges from $0.375$ to $0.550$.
This variability of saliency values degrades interpretability
because it makes \emph{thresholding} saliency value difficult:
There is no clear threshold to decide which pixel is positively important,
which is negatively important, and which is irrelevant.

To address this problem, we tweak the definition of RISE so that
the saliency value of an irrelevant pixel to be close to $0$,
a positively-important pixel to be positive,
and a negatively-important pixel to be negative;
in other words, our method generates a \emph{debiased} saliency map.
Concretely, we generate a saliency map using the following definition instead of Equation~(\ref{eq:riseFormula}):
\begin{eqnarray}
\label{eq:pnrise_def}
  S^{\mathrm{PN}}_{i,l}(\lambda) &:=&
    \mathbb{E}_{m}[M(i \odot m, l)|m(\lambda)=1]
    - \mathbb{E}_{m}[M(i \odot m, l)|m(\lambda)=0] ~.
\end{eqnarray}

Intuition behind Equation~(\ref{eq:pnrise_def}) is that we also need to use \emph{negative} saliency
$\mathbb{E}_{m}[M(i \odot m, l)|m(\lambda)=0]$ of a pixel $\lambda$ for the classification result $l$
in computing a saliency map, not only using the positive saliency $\mathbb{E}_{m}[M(i \odot m, l)|m(\lambda)=1]$
as RISE does.  The expression $\mathbb{E}_{m}[M(i \odot m, l)|m(\lambda)=0]$ indeed expresses
the negative saliency of $\lambda$ because its value becomes high if \emph{masking out} the pixel $\lambda$
increases the model confidence on average over the masks.  To give the saliency as a single number,
we calculate the difference of the positive saliency and the negative saliency.
As a result, the sign of $S^{\mathrm{PN}}_{i,l}(\lambda)$ indicates whether positive or negative saliency is dominant at a given pixel $\lambda$.

Calculating the difference of the positive saliency and the negative saliency also makes the saliency value of an irrelevant pixel to be $0$.
We can prove that
a saliency map generated from Equation (\ref{eq:pnrise_def}) is equipped with the following property
(\iffull
for the proof, see Appendix~\ref{sec:supplement_proof_pnrise}%
\fi
):

\begin{proposition}[Irrelevant pixels]
  \label{prop:irrelevant}
  If, for all binary mask samples,  the output probability of class $l$ from a black-box classifier $M$ does not vary
  no matter whether a pixel $\lambda$ in the mask $m$ is retained or masked
  (i.e. $M(i \odot m, l)_{m(\lambda)=1} = M(i \odot m, l)_{m(\lambda)=0}$),
  then $S^{\mathrm{PN}}_{i,l}(\lambda) = 0$.
\end{proposition}

This property justifies the following interpretation of $S^{\mathrm{PN}}_{i,l}$: A pixel whose value of $S^{\mathrm{PN}}_{i,l}$ is close to $0$ has almost no effect on the classification result.
It is worth noting that (1) the sign of the value of $S^{\mathrm{PN}}_{i,l}(\lambda)$ carries information on
whether the pixel $\lambda$ positively/negatively significant for the classification of $i$ to $l$ and
(2) its absolute value expresses how significant $\lambda$ is for the classification.
In the original RISE, we can observe only that
$\lambda$ is more/less positively significant than another pixel $\lambda'$ in the image $i$
if the value of $S_{i,l}(\lambda)$ is larger/smaller than $S_{i,l}(\lambda')$.
In \S\ref{sec:exp_results_neg}, we empirically justify the above interpretation and argue that these properties render
the interpretation of our saliency map easier than the one of the original RISE.

Equation~(\ref{eq:pnrise_def}) can be rewritten to the following equation
that computes $S^{\mathrm{PN}}_{i,l}$ by a weighted sum over all masks $m$
(see
\iffull
Appendix~\ref{sec:supplement_derive_pnrise} for the detailed derivation):
\fi
\begin{eqnarray}
\label{eq:pnrise_sum}
  S^{\mathrm{PN}}_{i,l}(\lambda) &=&
  \sum_m \frac{m(\lambda)-p}{p(1-p)} M(i \odot m, l) P[X=m] ~,
\end{eqnarray}
where $p = P[X(\lambda)=1]$.
We approximate $S^{\mathrm{PN}}_{i,l}$ by Monte Carlo sampling
and express it by the following equation:
\begin{eqnarray}
\label{eq:pnrise_deriv}
  S^{\mathrm{PN}}_{i,l}(\lambda) & \approx &
    \frac{1}{N} \sum_n \frac{m_n(\lambda) - p}{p(1-p)} M(i \odot m_n, l) ~,
\end{eqnarray}
where $N$ is the number of mask samples and $\{m_n\}_{n=1}^N$ are randomly-sampled masks.
We remark that Equation~(\ref{eq:pnrise_deriv}) is obtained from Equation~(\ref{eq:rise_deriv}) simply by replacing
$\frac{m_n(\lambda)}{p}$ with $\frac{m_n(\lambda) - p}{p(1-p)}$;
therefore, we can easily implement our debiasing method by modifying the implementation of RISE.

\subsection{Multi-Colored RISE (MC-RISE)}
\label{sec:method_mcrise}

We next present MC-RISE, our extension to RISE that explains the color-wise saliency of each pixel.
The main idea of MC-RISE is to use a colored mask instead of a binary mask that RISE uses.
Then, we can compute the positive/negative saliency of the given color component of a pixel from the response of the model to the masked image.
The pseudocode for MC-RISE is presented
\iffull
in Appendix~\ref{sec:pseudocode}.
\fi

\subsubsection{Color Masks}

We explain the method to generate color masks and color-masked images in MC-RISE.
We fix a set of colors $c_1,\dots,c_K \in \mathbb{R}^3$ where each $c_i$ is
a 3-tuple of color values in the RGB colorspace.
We are to generate a saliency map for each $c_i$.
To this end, MC-RISE generates \emph{colored masks} instead of the binary masks that RISE uses.
The saliency map is computed from the change in the average model confidence
for the color-masked image in a similar way to RISE.
We will explain the details of the saliency-map computation later.

To generate a colored mask, MC-RISE first generates a low-resolution color mask $m_{\mathrm{low}}(\lambda, k)$
of the size $h \times w$ (which is smaller than the input image's size), where $\lambda$ is the position of a pixel
and $k$ is the index for masking color in the color set $\{c_k\}_k$.
The low-resolution color mask $m_{\mathrm{low}}(\lambda, k)$ is generated as follows:
\begin{enumerate}
  \item For each pixel, MC-RISE randomly decides whether it is masked or not with a masking probability $p_{\mathrm{mask}}$.
  \item For each pixel that is decided to be masked, MC-RISE chooses the color used to mask it from the color set $\{c_k\}_k$ with the uniform probabilities.
  \item Then, the color mask $m_{\mathrm{low}}(\lambda, k)$ is defined as follows:
  \begin{equation}
  \label{eq:mcrise_lowresmask}
    m_{\mathrm{low}}(\lambda, k) = \begin{cases}
      1 \text{ if $\lambda$ is masked with color $k$.} \\
      0 \text{ otherwise.} \\
    \end{cases}
  \end{equation}
\end{enumerate}

Then, $m_{\mathrm{low}}(\lambda, k)$ is converted to a high-resolution color mask $m^{\mathrm{c}}(\lambda, k)$ by
(1) resizing $m_{\mathrm{low}}(\lambda, k)$ to the size $H \times W$ of input images using bilinear interpolation and
(2) shifting the resized mask by a random number of pixels from $(0, 0)$ up to $(\lfloor H/h \rfloor, \lfloor W/w \rfloor)$ (i.e., the size of a low-resolution pixel).
We remark that the computation of $m^{\mathrm{c}}(\lambda, k)$ corresponds to the preprocessing of a mask in RISE
mentioned in Remark~\ref{remark:optimization}.

We also compute a non-masked pixel map $m^{(0)}(\lambda)$ by
\begin{equation}
\label{eq:mcrise_nonmasked}
  m^{(0)}(\lambda; m^c) = 1 - \sum_{k=1}^{K} m^c(\lambda, k) ~.
\end{equation}
$m^{(0)}(\lambda; m^c)$ is $1$ for non-masked pixels and $0$ for masked pixels.

From a color mask $m^c$ and an input image $i$, the color-masked image $i'(\lambda; m^c)$ is computed by
$i'(\lambda; m^{\mathrm{c}}) = i(\lambda)m^{(0)}(\lambda; m^c) + \sum_{k=1}^{K} c_k m^{\mathrm{c}}(\lambda, k)$.
Because of Equations (\ref{eq:mcrise_lowresmask}) and (\ref{eq:mcrise_nonmasked}),
this expression can be seen as the alpha blending of an input image and the images uniformly filled with the masking colors $\{c_k\}_k$.

In Fig.~\ref{fig:method}, an example of a color mask $m^{\mathrm{c}}(\lambda, k)$,
a non-masked pixel map $m^{(0)}(\lambda)$, and a color-masked image $i'(\lambda; m^{\mathrm{c}})$
are presented.

\subsubsection{Color-aware Saliency Map}

The saliency map of MC-RISE is defined by
\begin{eqnarray}
  \label{eq:mcrise_def}
  S^{\mathrm{MC}}_{i,l}(\lambda, k) &:=&  \mathbb{E}_{m^{\mathrm{c}} \sim \mathcal{M}_c}[M(i'(\lambda; m^{\mathrm{c}}))|m^{\mathrm{c}}(\lambda, k)=1]\nonumber \\
  && - \mathbb{E}_{m^{\mathrm{c}} \sim \mathcal{M}_c}[M(i'(\lambda; m^{\mathrm{c}}))|m^{(0)}(\lambda)=1] ~,
\end{eqnarray}
where $\mathcal{M}_c$ denotes the probability distribution of color masks induced by the above mask-generation procedure.
$S^{\mathrm{MC}}_{i,l}(\lambda, k)$ represents how sensitively the model's confidence in classifying an image $i$ to class $l$ responds
when the pixel $\lambda$ in an input image is masked by the color $c_k$.

The debiasing method of a saliency map discussed in Section~\ref{sec:method_debias}
is incorporated also into the definition of color saliency maps (Equation (\ref{eq:mcrise_def})),
where the baseline saliency map for non-masked pixels
($\mathbb{E}_{m^{\mathrm{c}} \sim \mathcal{M}_c}[M(i'(\lambda; m^{\mathrm{c}}))|m^{(0)}(\lambda)=1]$, corresponding to the effect of retaining the pixel $\lambda$)
is subtracted from the raw saliency maps ($\mathbb{E}_{m^{\mathrm{c}} \sim \mathcal{M}_c}[M(i'(\lambda; m^{\mathrm{c}}))|m^{\mathrm{c}}(\lambda, k)=1]$, corresponding to the effect of masking the pixel $\lambda$).

The formula to estimate a color saliency map with Monte Carlo sampling can be derived in a similar way to \S\ref{sec:method_debias}
(see
\iffull
Appendix~\ref{sec:supplement_derive_mcrise}
\fi
for the detailed derivation);
\begin{eqnarray}
 \label{eq:mcrise_approx}
  S^{\mathrm{MC}}_{i,l}(\lambda, k) &=& \mathbb{E}_{m^{\mathrm{c}} \sim \mathcal{M}_c} \left[
      \left( \frac{m^{\mathrm{c}}(\lambda, k)}{p_{\mathrm{mask}}/K} - \frac{m^{(0)}(\lambda)}{1-p_{\mathrm{mask}}} \right) M(i'(\lambda; m^{\mathrm{c}})) \right]
      \nonumber \\
  & \approx &  \frac{1}{N} \sum_n \left( \frac{K m^{\mathrm{c}}_n(\lambda, k)}{p_{\mathrm{mask}}} -  \frac{m^{(0)}_n(\lambda)}{1-p_{\mathrm{mask}}} \right) M(i'(\lambda; m^{\mathrm{c}}_n)) ~,
\end{eqnarray}
where $N$ is the number of mask sampling and $\{m^{\mathrm{c}}_n\}_{n=1}^N$ are color masks sampled from the distribution $\mathcal{M}_c$.
In our implementation, $\frac{K m^{\mathrm{c}}_n(\lambda, k)}{p_{\mathrm{mask}}}$ and $\frac{m^{(0)}_n(\lambda)}{1-p_{\mathrm{mask}}}$ in Equation (\ref{eq:mcrise_approx})
are separately accumulated using the same samples of color masks $\{m^{\mathrm{c}}_n\}_n$;
$S^{\mathrm{MC}}_{i,l}(\lambda, k)$ is computed by subtracting the weighted-sum of the second term (baseline map)
from that of the first term (raw saliency maps), as illustrated in Fig.~\ref{fig:method}.

\subsubsection{Interpretation of Color Saliency Maps}

Since the debiasing method of \S\ref{sec:method_debias} is incorporated into the definition of color saliency maps (Equation (\ref{eq:mcrise_def})),
we can interpret the sign of a color saliency map as a positive/negative effect on the decision given by the model
as we discussed in \S\ref{sec:method_debias}.
However, because the sign of a color saliency map represents the expected change in the model confidence
by \emph{color-masking} a pixel $\lambda$, the interpretation of the sign of a saliency value is
different from the debiased RISE in \S\ref{sec:method_debias},
in which a saliency value represents the effect of \emph{retaining} original pixels.
Therefore, we can interpret the color saliency map $S^{\mathrm{MC}}_{i,l}(\lambda, k)$ as follows:
\begin{itemize}
  \item If  $S^{\mathrm{MC}}_{i,l}(\lambda, k)$ has a positive or negative value \emph{for all $k$}, the model's output probability of class $l$ should increase or decrease by any kind of color mask at a pixel $\lambda$, respectively.
  This suggests that the original \emph{texture} at pixel $\lambda$ is an obstacle or an important feature to the model, respectively, and its overall color is not relevant to the model's decision.

  \item If $S^{\mathrm{MC}}_{i,l}(\lambda, k) \sim 0$ \emph{for all $k$}, any kind of masking at the pixel $\lambda$ should not affect the model's output.
  It means that the pixel $\lambda$ is irrelevant to the model's decision.

  \item If  $S^{\mathrm{MC}}_{i,l}(\lambda, k) > 0$ \emph{for some $k$}, the model's output probability of class $l$ should increase by masking the pixel $\lambda$ with the color $c_k$.
  This indicates that the presence of the color $c_k$ at the pixel $\lambda$ is an important feature for the model to classify an image as class $l$, but the original image lacks this feature.

  \item If $S^{\mathrm{MC}}_{i,l}(\lambda, k) < 0$ \emph{for some $k$}, the model's output probability of class $l$ should decrease by masking the pixel $\lambda$ with the color $c_k$.
  This suggests that the pixel $\lambda$'s original color is an influential feature for the model to classify an image as class $l$ and masking it with the color $c_k$ degrade the model's confidence for class $l$.

  \item If $S^{\mathrm{MC}}_{i,l}(\lambda, k) \sim 0$ \emph{for some $k$},  color-masking the pixel $\lambda$ with the color $c_k$ should not affect the model's output.
  It suggests that the color $c_k$ is similar to the original color of the pixel $\lambda$ from the viewpoint of the model, therefore masking $\lambda$ with other colors would affect the model's output.

\end{itemize}

\section{Experiments}
\label{sec:exp}

\subsection{Experimental Settings}
\label{sec:exp_details}

We use LIME~\cite{lime} and RISE~\cite{DBLP:conf/bmvc/PetsiukDS18} as the baseline for the evaluation of our methods.
We designated two kinds of models trained with different datasets for the evaluation.
\begin{itemize}
  \item Models trained with GTSRB~\cite{gtsrb} dataset: We trained
  (1) VGG-16~\cite{vgg} model with batch normalization and (2) ResNet-50~\cite{resnet} model.
  We trained the models for $90$ epochs with momentum SGD using cross-entropy loss.
  We set the learning rate to $10^{-1}$, the momentum to $0.9$, and the weight decay to $5\times 10^{-4}$;
  we decayed the learning rate by a factor of $0.1$ in every $30$ epochs.
  We resized images to $H \times W = 96 \times 96$ pixels for training and evaluation.
  \item Models trained with ImageNet\cite{imagenet} dataset:
  We used the pretrained ResNet-50 model provided by PyTorch~\cite{paszke_pytorch_2019} for the evaluation.
  We cropped and resized images to $H \times W = 224 \times 224$ pixels for the evaluation.
\end{itemize}

For MC-RISE and RISE, the number of samples $N$ was set to $8000$ and
the masking probability was set to $0.5$.
For MC-RISE, we used the following five colors:
red ($c_k = (255,0,0)$ in the RGB colorspace), green ($c_k = (0, 255, 0)$), blue ($c_k = (0, 0, 255)$),
white ($c_k = (255, 255, 255)$), and black ($c_k = (0, 0, 0)$).
For LIME, the parameters that we used were the same as the original paper~\cite{lime}.

We also applied MC-RISE to a person re-identification model with metric learning.
See
\iffull
Appendix~\ref{sec:reid}
\fi
for details.

\subsection{Results}

\subsubsection{Qualitative Evaluation of the Debiasing Method}
\label{sec:exp_results_neg}

\begin{figure}[tb]
  \centering
  \includegraphics[keepaspectratio, width=0.60\linewidth]{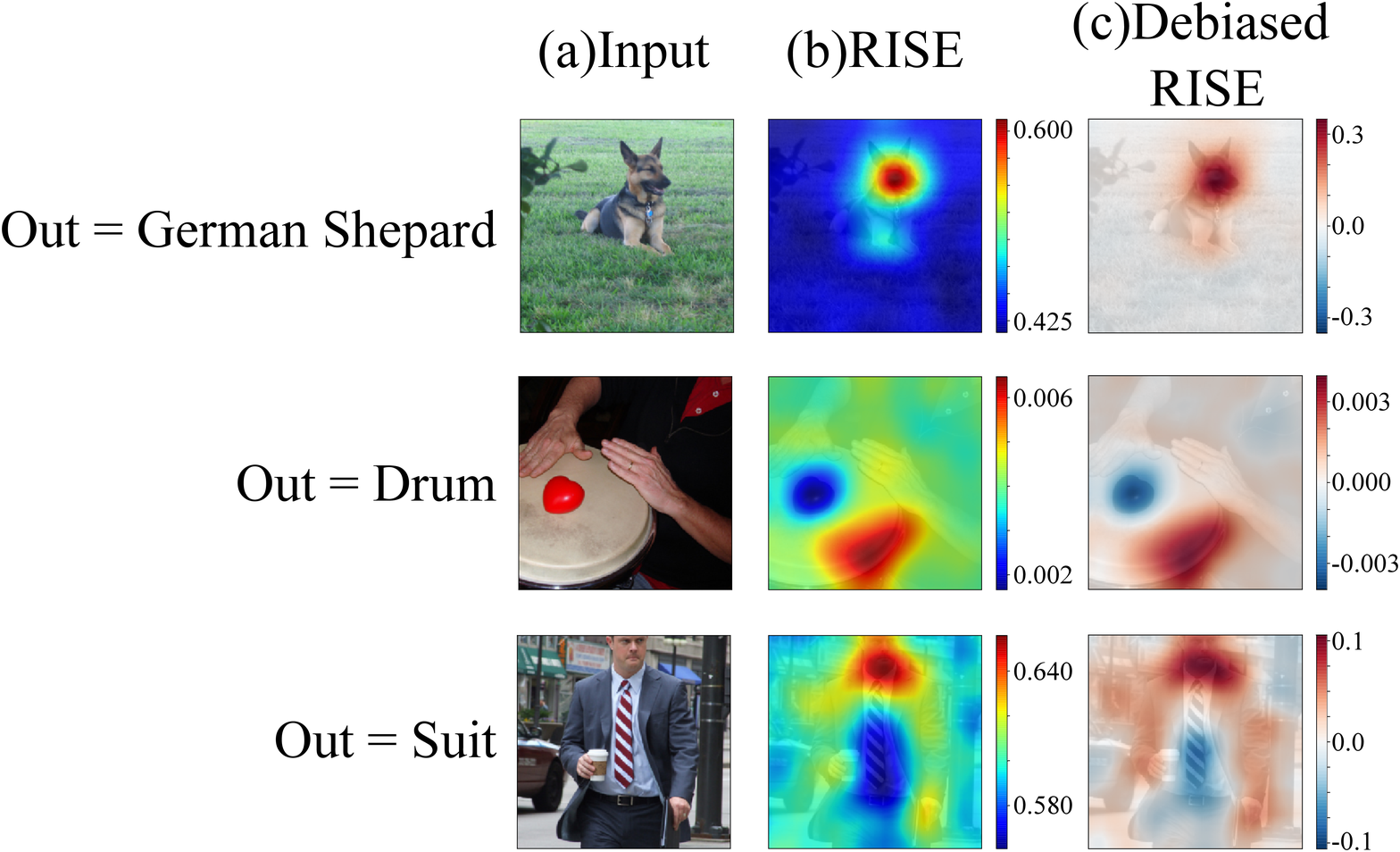}
  \caption{Visual comparison between (b)raw saliency maps of RISE\cite{DBLP:conf/bmvc/PetsiukDS18} and (c)debiased saliency maps calculated by our method. The parts of the image contributing positively/negatively to the output class have positive/negative intensity in (c), respectively. The background regions irrelevant to the model's output have near-zero intensity in (c). Best viewed in color.}
  \label{fig:pnrise}
\end{figure}

We qualitatively evaluated the debiasing method presented in \S\ref{sec:method_debias}.
Fig.~\ref{fig:pnrise} compares the saliency maps of RISE and our debiased saliency map;
we used ImageNet dataset to generate these maps.
In the saliency map generated by the original RISE (Fig.~\ref{fig:pnrise}(b)),
the saliency values of the background regions---which should be irrelevant to the decision of the model---largely differs depending on the input images
both in their absolute values and in their relative magnitude
(i.e., pseudo-color in the plots).
In our debiased maps (Fig.~\ref{fig:pnrise}(c)), the background regions of all the images
have the saliency values that are close to zero.
These results demonstrate that the debiased RISE computes a saliency map
in which the saliency values of the irrelevant regions are
close to zero as we intend.

The saliency maps on the second and third rows in Fig.~\ref{fig:pnrise}(c) shows that
the saliency values near the objects of the output class (i.e., Drum and Suit) are positive
and that the saliency values near the ``noisy'' objects (i.e., heart-shaped weight and necktie)
are negative.  These behaviors are also in accordance with the interpretation we presented
in \S\ref{sec:method_debias}.

\subsubsection{Qualitative Evaluation of MC-RISE}

\begin{figure}[tb]
  \centering
  \includegraphics[keepaspectratio, width=\linewidth]{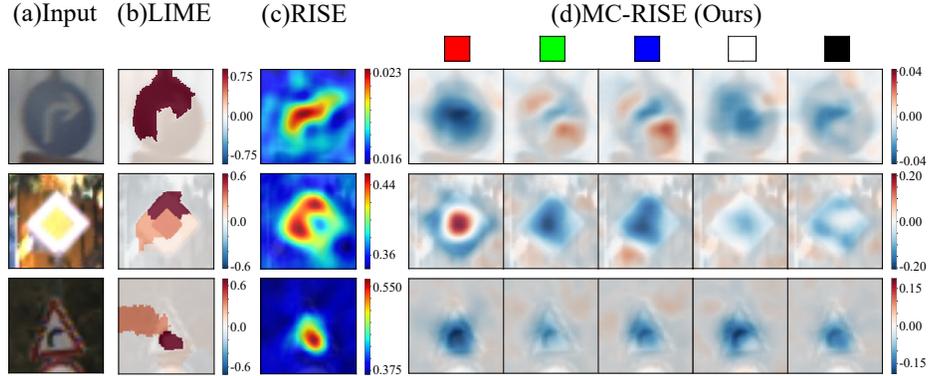}
  \caption{Visual comparison among the saliency maps of (b) LIME~\cite{lime}, (c) RISE~\cite{DBLP:conf/bmvc/PetsiukDS18}, and (d) the color saliency maps of MC-RISE in the GTSRB dataset. The boxes in the top row in (d) present the masking color used to obtain the maps in the same column. All samples are correctly classified by the model. Best viewed in color.}
  \label{fig:mcrise_gtsrb}
\end{figure}

Fig.~\ref{fig:mcrise_gtsrb} compares saliency maps generated by LIME~\cite{lime},
RISE~\cite{DBLP:conf/bmvc/PetsiukDS18}, and our method MC-RISE; we used GTSRB dataset to generate these saliency maps.
The saliency values of MC-RISE on the third row are not sensitive to
the colors (i.e. the sign and the values of $S^{\mathrm{MC}}_{i,l}(\lambda, k)$ is similar for all $k$).
As we mentioned in \S\ref{sec:method_mcrise},
this suggests that the texture information is important for the model
than the color information.

The saliency map of the image on the first row in Fig.~\ref{fig:mcrise_error}
is sensitive to colors.  In this image, the saliency near the white arrow
is negative for all the colors; this suggests that the texture of this region (i.e., the arrow shape)
is important for the model.  In contrast, the values near the peripheral region of the sign
in the green and the blue saliency maps are positive; this suggests that
the blue-like color of this region is important for the model's decision.
Notice that LIME nor RISE can detect the sensitivity to colors;
in fact, these parts identified as salient in the green and the blue saliency maps
are not necessarily identifiable in the saliency maps generated by LIME and RISE.

\begin{figure}[tb]
  \centering
  \includegraphics[keepaspectratio, width=\linewidth]{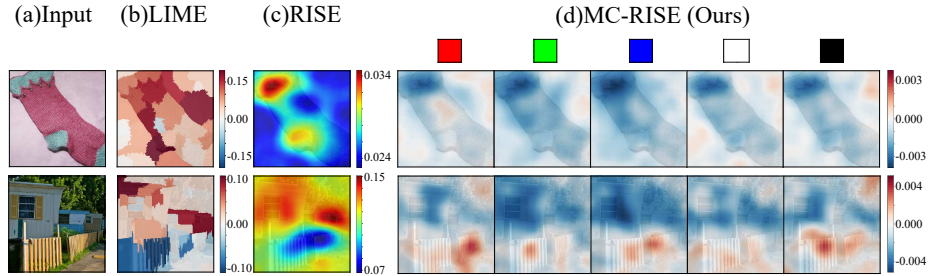}
  \caption{Visual comparison among the saliency maps of (b) LIME~\cite{lime}, (c) RISE~\cite{DBLP:conf/bmvc/PetsiukDS18}, and (d)the color saliency maps of MC-RISE in the ImageNet dataset. All samples are correctly classified by the model. Best viewed in color.}
  \label{fig:mcrise_imagenet}
\end{figure}

Fig. \ref{fig:mcrise_imagenet} shows the comparison of the saliency maps in ImageNet dataset.
Compared to maps in the GTSRB dataset, many of the maps in the ImageNet dataset have
low color-dependency among the color saliency maps.
This suggests that the model's decision is more dependent on
the detailed texture of objects than on the color of image regions.
In such cases, we can
extract information about positive or negative saliency from the color saliency maps.
For example, in the third-row samples in Fig. \ref{fig:mcrise_imagenet},
color saliency maps are interpreted to mean that the upper part of
the image is important for the classification output (``Mobile home'' class),
but the fences in the lower part of the image are reducing the model's confidence for the output class.
From these observations, it is expected that MC-RISE can extract more useful information
in domains where the color in image regions is highly important for classification, such as the GTSRB dataset,
than domains where detailed texture is important such as the ImageNet dataset.

\subsubsection{Quantitative Evaluation of MC-RISE}

\begin{table}[tb]
  \centering
  \caption{Evaluations of CA-deletion metric (lower is better) in LIME~\cite{lime}, RISE~\cite{DBLP:conf/bmvc/PetsiukDS18}, and MC-RISE (ours).}
  \label{tab:mcrise_delins}%

  \begin{tabular}{c|c|ccc}
    \hline
    Dataset & Model & LIME  & RISE  & MC-RISE \bigstrut\\
    \hline
    \hline
    \multirow{2}[2]{*}{GTSRB} & VGG-16 & 0.1324 & 0.0664 & \textbf{0.0270} \bigstrut[t]\\
          & ResNet50 & 0.1294 & 0.0627 & \textbf{0.0204} \bigstrut[b]\\
    \hline
    ImageNet & ResNet50 & 0.1146 & 0.1046 & \textbf{0.0980} \bigstrut\\
    \hline
  \end{tabular}%

\end{table}

For the quantitative evaluation of MC-RISE, we first would like to establish the evaluation metrics we use.
We adapt the \emph{deletion} metric proposed by Petsiuk et al.~\cite{DBLP:conf/bmvc/PetsiukDS18}---which measures how well a saliency-map--based explanation of an image classification result localizes the important pixels---to our color-aware setting.

To explain the deletion metric, let $i$ be an image and $M$ be a model.
The deletion metric is the AUC value of the plot of $f$ where $f(j)$
is the model confidence of the image in which the 1st to the $j$-th important pixels
are removed.
The lower the deletion metric is, the better the saliency map localizes the important region;
a lower metric value means that the model confidence drops rapidly because the saliency map successfully
points out the important pixels.

Our metric, called \emph{CA-deletion (color-aware deletion)}
is derived from the deletion metric as follows.
In the CA-deletion metric, pixels in an input image are removed by masking them with \emph{the most sensitive color}
obtained from the color saliency maps.  Concretely, pixel $\lambda$ is removed in the ascending order of
$\mathrm{min}_k[S^{\mathrm{MC}}_{i,l}(\lambda, k)]$ and the removed pixel is filled by the color $c_k$
where $k = \mathrm{argmin}_k[S^{\mathrm{MC}}_{i,l}(\lambda, k)]$.
The model confidence for the pixel-removed images are plotted against the pixel-removal fractions;
the CA-deletion value is the area-under-curve (AUC) of the plot.
If a color saliency map correctly captures the model's sensitivity to color masking,
this removal method must be effective to reduce the model confidence,
resulting in a low value of the CA-deletion metric.\footnote{Notice that other common metrics
(e.g., pointing game) are not appropriate for evaluating MC-RISE because these metrics
focus on the \emph{positional} saliency; they do not consider the color saliency.}

Table \ref{tab:mcrise_delins} shows the evaluation result of MC-RISE in comparison with LIME and RISE.
MC-RISE consistently outperforms RISE and LIME especially by a large margin in the GTSRB dataset.
This indicates that MC-RISE can correctly capture the color-dependent sensitivity of the model,
and effectively localize important pixels in an input image by using this information.
The large margin in the GTSRB dataset is in accordance with our expectation that MC-RISE is
especially effective in a domain, such as traffic sign classification, where color information conveys important information.

\subsubsection{Use Case: Error Analysis Using MC-RISE}

\begin{figure}[tb]
  \centering
  \includegraphics[keepaspectratio, width=\linewidth]{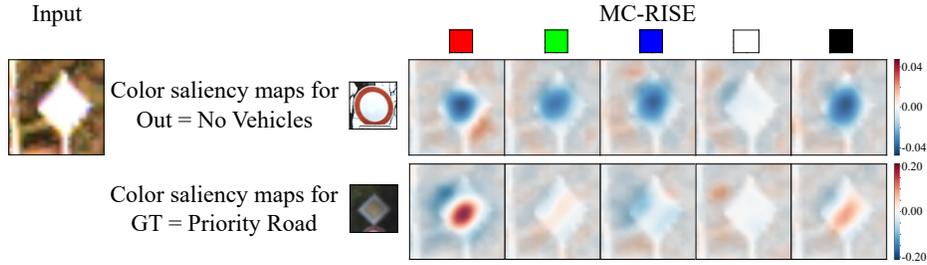}
  \caption{Error analysis example in the GTSRB classification model with MC-RISE.
    Sensitivities for a red color mask at the center of the sign are opposite for the output class (No Vehicle) and the GT class (Priority road). It suggests that the absence of red color at the sign center is the cause of the error. Best viewed in color.}
  \label{fig:mcrise_error}
\end{figure}

We demonstrate a use case of MC-RISE for analyzing misclassified samples.
Fig.~\ref{fig:mcrise_error} shows the color saliency maps generated by MC-RISE
for a sample in GTSRB dataset that is misclassified by our model.
The input is an image of a ``Priority road'' sign, whereas our model misclassified it as a ``No vehicles'' sign.

The saliency maps for the output ``No vehicles'' class show that the colors other than white
at the center part of the sign have a negative effect in classifying this image as ``No vehicles''.
This suggests that, if the color of the center part were not white, then this image would not be classified as ``No vehicles'' class.

In contrast, in the saliency maps of the GT class (i.e., ``Priority road'' sign),
the saliency map for red color has a strong positive value at the center of the sign,
suggesting that the confidence of ``Priority Road'' class would largely increase
if the color of the central part were red.
From the above observations, we can hypothesize that the cause of the error is
the absence of red color at the center of the sign (possibly due to overexposure).
As one can observe, the information obtained by MC-RISE in addition to that obtained by the other methods, such as color-sensitivity or negative saliency,
is indeed useful to explore the cause of the misclassification,
especially when there is no objects which clearly hinder the correct decision of the model in an input image.

\section{Conclusion}

\label{sec:conclusion}

We proposed two extensions of a model-agnostic interpretation method RISE~\cite{DBLP:conf/bmvc/PetsiukDS18}.
One is debiased RISE; it computes a saliency map in which the saliency of an unimportant pixel is close to $0$;
this makes the interpretation of a saliency map easier than RISE.
Another is MC-RISE, which computes color-wise saliency;
a saliency map for a color $c$ computed by MC-RISE explains the saliency of the color $c$ in each pixel.
Such color-wise saliency is useful in application domains such as road sign recognition, in which colors
convey significant information.  We empirically demonstrated the effectiveness of our extensions.

We plan to extend our idea to use other features than colors;
for example, by using masks that blur a part of an image, we expect
that we can generate a saliency map that explains the significance of the clarity of the region.

\subsubsection*{Acknowledgment}
We thank the reviewers for their fruitful comments.



\bibliographystyle{splncs}
\bibliography{main}


\iffull
\appendix
\section{Derivation of Equation (\ref{eq:pnrise_sum})}
\label{sec:supplement_derive_pnrise}

We rewrite the definition of $S^{\mathrm{PN}}_{i,l}(\lambda)$ (Equation (\ref{eq:pnrise_def})) as the sum over all possible masks $m$;
\begin{eqnarray}
\label{eq:s_pn1}
S^{\mathrm{PN}}_{i,l}(\lambda) &=&
\mathbb{E}_{m}[M(i \odot m, l)|m(\lambda)=1]
\nonumber
- \mathbb{E}_{m}[M(i \odot m, l)|m(\lambda)=0] \nonumber \\
  &=&  \sum_m M(I \odot m, l) \nonumber \\
  && \times \left( P[X=m|X(\lambda)=1] - P[X=m|X(\lambda)=0] \right) ~.
\end{eqnarray}
$P[X=m|X(\lambda)=1]$ and $P[X=m|X(\lambda)=0]$ are expressed as
\begin{eqnarray}
\label{eq:m_one}
P[X=m|X(\lambda)=1] &=& \frac{P[X=m, X(\lambda)=1]}{P[X(\lambda)=1]} \nonumber \\
&=& \begin{cases}
  0 &\text{ if } m(\lambda) = 0 \\
  \frac{P[X=m]}{P[X(\lambda)=1]} &\text{ if } m(\lambda) = 1
\end{cases} \nonumber \\
&=& \frac{m(\lambda)P[X=m]}{p} ~,
\end{eqnarray}
\begin{eqnarray}
\label{eq:m_zero}
P[X=m|X(\lambda)=0]
&=& \begin{cases}
\frac{P[X=m]}{P[X(\lambda)=0]} &\text{ if } m(\lambda) = 0 \\
0 &\text{ if } m(\lambda) = 1
\end{cases} \nonumber \\
&=& \frac{(1-m(\lambda))P[X=m]}{1-p} ~,
\end{eqnarray}
where $p = P[X(\lambda)=1] = 1 - P[X(\lambda)=0]$.
Substituting these expressions in the expression of $S^{\mathrm{PN}}_{i,l}(\lambda)$ (Equaiton (\ref{eq:s_pn1})), we obtain Equation (\ref{eq:pnrise_sum});
\begin{eqnarray}
S^{\mathrm{PN}}_{i,l}(\lambda)
  &=& \sum_m \left(
    \frac{m(\lambda)}{p} - \frac{1-m(\lambda)}{1-p} \right) M(I \odot m, l) P[X=m] \nonumber \\
&=&   \sum_m \frac{m(\lambda)-p}{p(1-p)} M(i \odot m, l) P[X=m] ~.
\end{eqnarray}

\section{Derivation of Equation (\ref{eq:mcrise_approx})}
\label{sec:supplement_derive_mcrise}

Following the similar derivation in \S\ref{sec:supplement_derive_pnrise}, the definition of $S^{\mathrm{MC}}_{i,l}(\lambda)$ (Equation (\ref{eq:mcrise_def})) can be expressed as
\begin{eqnarray}
\label{eq:s_mc1}
S^{\mathrm{MC}}_{i,l}(\lambda, k) &:=&  \mathbb{E}_{m^{\mathrm{c}} \sim \mathcal{M}_c}[M(i'(\lambda; m^{\mathrm{c}}))|m^{\mathrm{c}}(\lambda, k)=1] \nonumber \\
&& - \mathbb{E}_{m^{\mathrm{c}} \sim \mathcal{M}_c}[M(i'(\lambda; m^{\mathrm{c}}))|m^{(0)}(\lambda)=1] \nonumber \\
&=&  \sum_m M(i'(\lambda; m^{\mathrm{c}}), l) \nonumber \\
&&\times \left( P[X=m^{\mathrm{c}}|X(\lambda, k)=1]
- P[X=m^{\mathrm{c}}|X^{(0)}(\lambda)=1] \right) \nonumber \\
&=& \sum_m \left(
\frac{m^{\mathrm{c}}(\lambda, k)}{P[X(\lambda, k)=1]} - \frac{m^{(0)}(\lambda)}{P[X^{(0)}(\lambda)=1]]} \right) \nonumber \\
&& \times M(i'(\lambda; m^{\mathrm{c}}), l) P[X=m] ~.
\end{eqnarray}
In the mask generation process, whether a pixel is masked or retained is randomly determined with the masking probability $p_{\mathrm{mask}}$, and for a masked pixel, the masking color is sampled from the uniform distribution over $K$ colors;
hence, $P[X(\lambda, k)=1] = p_{\mathrm{mask}}\frac{1}{K}$ and $P[X^{(0)}(\lambda)=1]= 1-p_{\mathrm{mask}}$.
Substituting them in the expression of $S^{\mathrm{MC}}_{i,l}(\lambda)$ (Equation (\ref{eq:s_mc1})) yields Equation (\ref{eq:mcrise_approx});
\begin{eqnarray}
  S^{\mathrm{MC}}_{i,l}(\lambda, k) &=& \mathbb{E}_{m^{\mathrm{c}} \sim \mathcal{M}_c} \left[
      \left( \frac{m^{\mathrm{c}}(\lambda, k)}{p_{\mathrm{mask}}/K} - \frac{m^{(0)}(\lambda)}{1-p_{\mathrm{mask}}} \right) M(i'(\lambda; m^{\mathrm{c}})) \right] ~.
\end{eqnarray}

\section{Proof of Proposition \ref{prop:irrelevant}}
\label{sec:supplement_proof_pnrise}

Let $\lambda \in \Lambda$ be a fixed pixel and $A$ be the set of all possible masks.
We define a disjoint partition of $A$ by  $A^{+}=\{m \in A | m(\lambda)=1\}$ and $A^{-}=\{m \in A | m(\lambda)=0\}$.
From the definition of $S^{\mathrm{PN}}_{i,l}(\lambda)$ (Equation (\ref{eq:pnrise_def})), we get
\begin{eqnarray}
\label{eq:proof_1}
S^{\mathrm{PN}}_{i,l}(\lambda) &=&
\sum_{m \in A^{+}} M(I \odot m, l) P[X=m|X(\lambda)=1] \nonumber \\
&& - \sum_{m \in A^{-}} M(I \odot m, l) P[X=m|X(\lambda)=0]
\end{eqnarray}
because $P[X=m|X(\lambda)=1] = 0$ if $m(\lambda)=0$ and vice versa.

Let $F_{\lambda}: A \rightarrow A$ be the function which flips the mask value at pixel $\lambda$.
$F_{\lambda}$ induces a one-to-one correspondence between the masks in $A^{+}$ and $A^{-}$.
Therefore, Equation (\ref{eq:proof_1}) is expressed as
\begin{eqnarray}
\label{eq:proof_2}
S^{\mathrm{PN}}_{i,l}(\lambda) &=&
\sum_{m \in A^{+}} M(I \odot m, l) P[X=m|X(\lambda)=1] \nonumber \\
&& - \sum_{m \in A^{+}} M(I \odot F_{\lambda}(m), l) P[X=F_{\lambda}(m)|X(\lambda)=0] ~.
\end{eqnarray}

We can rewrite $P[X=F_{\lambda}(m)|X(\lambda)=0]$ as
\begin{eqnarray}
P[X=F_{\lambda}(m)|X(\lambda)=0] &=& \delta_{F_{\lambda}(m)(\lambda), 0} \prod_{\kappa \in \Lambda \setminus \{\lambda\}}p^{m(\kappa)}(1-p)^{1-m(\kappa)} \nonumber \\
&=& \delta_{m(\lambda), 1} \prod_{\kappa \in \Lambda \setminus \{\lambda\}}p^{m(\kappa)}(1-p)^{1-m(\kappa)} \nonumber \\
&=& P[X=m|X(\lambda)=1] ~,
\end{eqnarray}
where $\delta_{i,j}$ is the Kronecker delta and $p$ is the masking probability for a pixel. Therefore, Equation (\ref{eq:proof_2}) is rewritten as
\begin{eqnarray}
\label{eq:proof_3}
S^{\mathrm{PN}}_{i,l}(\lambda)
&=&
\sum_{m \in A^{+}} \left\{ M(I \odot m, l) - M(I \odot F_{\lambda}(m), l) \right\} \nonumber \\
&&\times P[X=m|X(\lambda)=1] ~.
\end{eqnarray}

Since the premise in Proposition~\ref{prop:irrelevant} is expressed as
\begin{equation}
M(I \odot m, l) = M(I \odot F_{\lambda}(m), l) \text{ for all } m \in A^{+} ~,
\end{equation}
we obtain $S^{\mathrm{PN}}_{i,l}(\lambda) = 0$.

\section{Additional Experiments with Person ReID model}
\label{sec:reid}

This section presents the application of MC-RISE to
a metric-learning-based person re-identification (ReID) model.

A metric-learning-based person ReID model
takes an image $i$ of a person as input and outputs
the feature vector $v_i$ for the input image.
The model is trained so that, if it is given a pair of images
$i$ and $i'$, then the distance between $v_i$ and $v_{i'}$ is small
if $i$ and $i'$ are likely to be the images of the same person.
We designate a set of gallery images; at inference time,
we compare the feature vector of a query image with those of the gallery images
and retrieve a gallery image that belongs to the same person as the query.

We adapted MC-RISE as follows to apply it to a person ReID model.
\begin{itemize}
  \item Unlike a standard classification task where an image belongs to a unique class,
  a person ReID task has multiple correct gallery images for one query image in general.
  In this experiment, we only consider the gallery image of the top-1 match to a query image $i$ as the correct label;
  the feature distance to the top-1 match image is used as the output of the black-box model $M(i,l)$,
  that is used by MC-RISE.
  Hence, MC-RISE visualizes how the \emph{similarity} to the top-1 match image responds to the color masking of an input image.
  \item As for the definition of feature distance, the simple Euclidean distance between feature vectors is not appropriate
  because the weighted sum of color masks would be dominated by outlier samples with large feature distance,
  resulting in the uninterpretable saliency maps.
  We computed $M(i,l)$ by the following formula:
  \begin{equation}
    \label{eq:preid_output}
    M(i,l) = \exp\left(-\frac{d(f_{i}, f_{\mathrm{match}})}{d_0}\right) .
  \end{equation}
  Here, $d(f_{i}, f_{\mathrm{match}})$ is the feature distance between an input image $i$
  and the top-1 match image, and $d_0$ is a typical scale of the distance,
  for which we used the feature distance between the original query image and
  the top-1 match image.
  The resulting saliency maps are not much affected by the outliers
  since Equation~(\ref{eq:preid_output}) becomes nearly zero for an outlier with a large distance.
\end{itemize}

In our experiment, we visualized the color saliency maps for the Market-1501 dataset~\cite{zheng_market1501_2015}.
For the evaluation, we used the pretrained OSNet($\times 1.0$)~\cite{zhou_osnet_2019} model
provided by Torchreid library~\cite{torchreid}.
The parameters for MC-RISE were the same as \S\ref{sec:exp_details}
except that the masking probability was set to $0.1$.

\begin{figure}[tb]
  \centering
  \includegraphics[keepaspectratio, width=\linewidth]{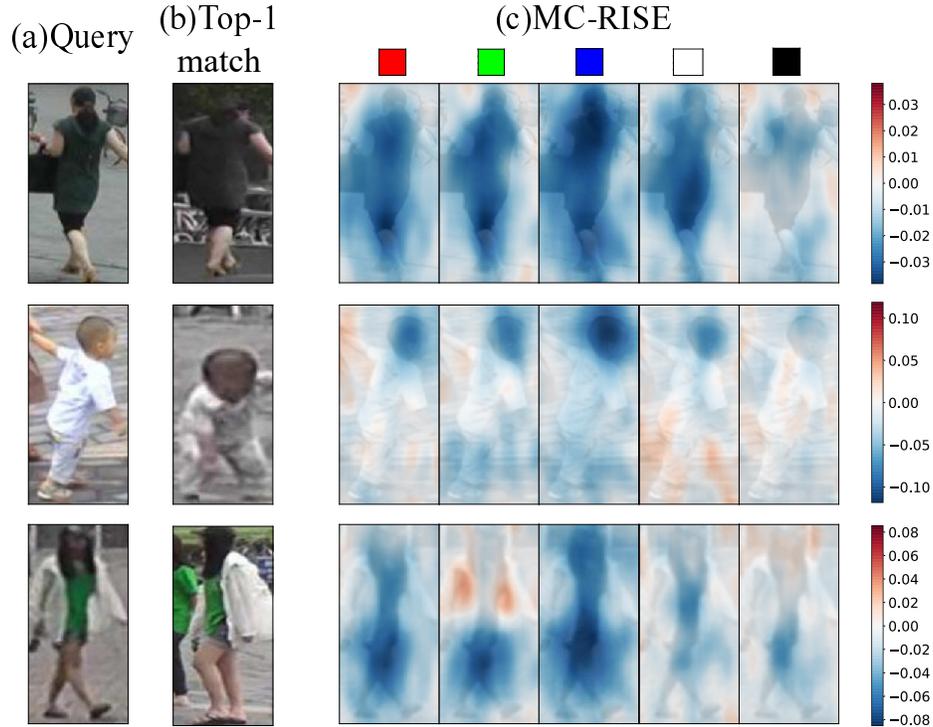}
  \caption{The color saliency maps generated by MC-RISE for the Market-1501 dataset.
  The saliency maps in (c) visualize the color-wise saliency for the similarity between (a) a query image
  and (b) the top-1 match image in gallery images.
  All queries are correctly matched by the model.  Best viewed in color.}
  \label{fig:mcrise_preid}
\end{figure}

Fig.~\ref{fig:mcrise_preid} shows the saliency maps generated for the Market-1501 dataset.
For most of the query images, the saliency maps have negative values on the entire body
as the top row sample shows.
This suggests that the model compares the whole parts of the body in a query image with that of the gallery images;
if the colors of corresponding parts disagree, it largely diminishes the similarity between images.
However, for some queries such as the ones in the middle or the bottom row,
the saliency maps indicate that the model pays close attention to
a specific part of the body (e.g., head in the middle row sample)
or the specific color of clothing (e.g., green color clothing in the bottom row sample).
These results demonstrate that MC-RISE can also be applied to metric learning problems,
such as person ReID task, and can visualize the characteristics of the model's decision.

\section{MC-RISE with $K=8$}

\begin{figure}[tb]
  \centering
  \includegraphics[keepaspectratio, width=\linewidth]{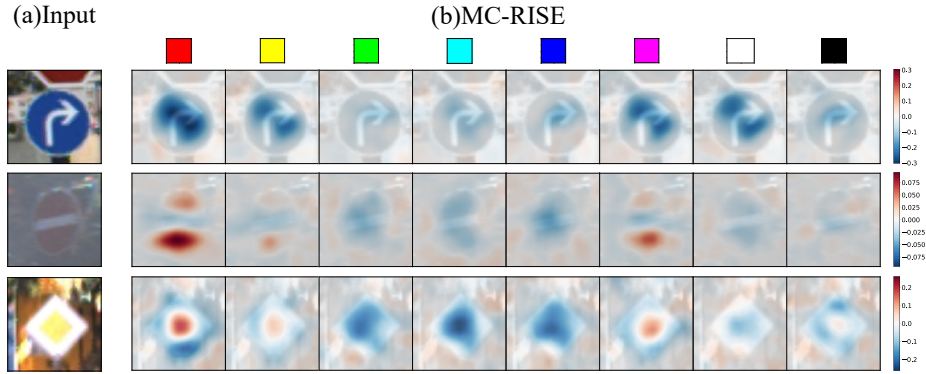}
  \caption{Saliency maps generated by MC-RISE with $K=8$ for GTSRB dataset.}
  \label{fig:mcrise_8color}
\end{figure}

In the experiments with GTSRB dataset in \S\ref{sec:exp},
we applied MC-RISE by setting the number of colors to $5$ (i.e., $K=5$).
Fig.~\ref{fig:mcrise_8color} shows several saliency maps generated by MC-RISE
wherein we set $K$ to $8$; the other settings are kept the same as in \S\ref{sec:exp}.

Although the tendency of the saliency maps is by and large the same as that in \S\ref{sec:exp},
it is worth noting that, by using more colors,
we can read out more information from
the saliency map in the bottom row in Fig.~\ref{fig:mcrise_8color}
than that in Fig.~\ref{fig:mcrise_gtsrb} in \S\ref{sec:exp}.
We can observe that, in addition to red at the center of the sign,
stronger yellow and magenta at the center would make the confidence more solid.

\section{Pseudocode of MC-RISE}
\label{sec:pseudocode}

The pseudocode of MC-RISE is presented in Algorithm~\ref{algo:mcrise}.

\begin{algorithm}[tb]
  \DontPrintSemicolon

  \caption{MC-RISE algorithm}
  \label{algo:mcrise}

  \KwInput{Input image $i(\lambda)$, Target label $l$, Black-box model $M$,
      Color set $\{c_k\}_{k=1}^K$, The number of masks $N$, Low-resolution mask size $h \times w$}
  \KwOutput{Color saliency maps $S^{\mathrm{MC}}_{i,l}(\lambda, k)$}
    $S^{\mathrm{raw}}(\lambda, k) \leftarrow 0$, $S^{\mathrm{baseline}}(\lambda) \leftarrow 0$ for all $\lambda, k$\;
    \For{$n=1$ to $N$}
    {
      \tcp{generating color masks}
      $m_{\mathrm{mask}} \leftarrow$ randomly sample $h \times w$ binary mask with the masking probability $p_{mask}$. \;
      \For{pixel $\lambda$ in $h \times w$ image}
      {
          \tcp{eq.(6)}
          $m_{\mathrm{low}}(\lambda, k) \leftarrow 0$ for all $k=1 \ldots K$\;
          \If{$m_{\mathrm{mask}}(\lambda) = 1$}
          {
              $k' \leftarrow$ randomly sample the index of masking color from $\{1 \ldots K\}$ \;
              $m_{\mathrm{low}}(\lambda, k') \leftarrow 1$ \;
          }
      }

      $m^c_n \leftarrow \mathit{bilinear\_interpolation}(m_{\mathrm{low}})$ \;
      $m^c_n \leftarrow \mathit{random\_shift}(m^c_n)$ \;
      $m^{(0)}_n(\lambda) \leftarrow 1 - \sum_{k=1}^{K} m^c_n(\lambda, k)$ for all $\lambda$ \tcp*{eq.(7)}

      \tcp{computing saliency maps; eq.(10)}
      $i'(\lambda) \leftarrow i(\lambda)m^{(0)}_n(\lambda) + \sum_{k=1}^{K} c_k m^{\mathrm{c}}_n(\lambda, k)$ for all $\lambda$ \tcp*{eq.(8)}
      $p_{\mathrm{out}} \leftarrow M(i',l)$ \;
      $S^{\mathrm{raw}}(\lambda, k) \mathrel{+}= \frac{K m^c_n(\lambda, k)}{p_{\mathrm{mask}}}p_{\mathrm{out}} $ for all $\lambda, k$ \;
      $S^{\mathrm{baseline}}(\lambda) \mathrel{+}= \frac{m^{(0)}_n(\lambda)}{(1 - p_{\mathrm{mask}})} p_{\mathrm{out}}$  for all $\lambda$ \;
    }
    $S^{\mathrm{MC}}_{i,l}(\lambda, k) \leftarrow (S^{\mathrm{raw}}(\lambda, k) - S^{\mathrm{baseline}}(\lambda)) / N$ for all $\lambda, k$ \;

\end{algorithm}



\else
\fi

\end{document}